\documentclass{vgtc}                          

\ifpdf
  \pdfoutput=1\relax                   
  \usepackage{graphicx}                
  \DeclareGraphicsExtensions{.pdf,.png,.jpg,.jpeg} 
\else
  \usepackage{graphicx}                
  \DeclareGraphicsExtensions{.eps}     
\fi%

\graphicspath{{figures/}{pictures/}{images/}{./}} 

\usepackage{mathptmx}
\usepackage{epstopdf}
\usepackage{caption}
\usepackage{times}
\usepackage[lined,algonl,boxed]{algorithm2e}
\usepackage{amssymb,amsmath}
\usepackage{bbm}
\usepackage{multirow}
\usepackage{subfigure}
\usepackage{paralist}
\usepackage{tikz}
\usepackage{multirow}
\usepackage{color}
\usepackage{setspace}
\usepackage{microtype,hyphenat,balance}
\usepackage{array}
\usepackage{url}
\usepackage{bm}
\usepackage{algorithm2e}
\usepackage{algpseudocode}
\usepackage{amsmath}
\usepackage{booktabs}
\usepackage[implicit=false]{hyperref}

\definecolor{weikaiColor}{RGB}{56,118,29}
\definecolor{darkgreen}{RGB}{0,128,0}
\newcommand{\liu}[1]{\textcolor{black}{#1}}
\newcommand{\shixia}[1]{\textcolor{black}{#1}}
\newcommand{\mc}[1]{\textcolor{black}{#1}}
\newcommand{\weikai}[1]{\textcolor{black}{#1}}
\newcommand{\yafeng}[1]{\textcolor{black}{#1}}
\newcommand{\lu}[1]{\textcolor{black}{#1}}
\newcommand{\ross}[1]{\textcolor{black}{#1}}
\newcommand{\mcr}[1]{\textcolor{black}{#1}}
\newcommand{\minorchange}[1]{\textcolor{black}{#1}}
\newcommand{\weikairevision}[1]{\textcolor{black}{#1}}

\onlineid{1087}

\vgtccategory{Research}

\title{Diagnosing Concept Drift \liu{with Visual Analytics}}
\author{
Weikai Yang\thanks{e-mail:\{\{ywk19, liz11, ckl17\}@mails., shixia@\}tsinghua.edu.cn. Weikai Yang and Zhen Li are joint first authors. Shixia Liu is the corresponding author.},
Zhen Li$^{*}$,
Mengchen Liu\thanks{e-mail:mengcliu@microsoft.com},
Yafeng Lu\thanks{e-mail:luyafeng.cn@gmail.com},
Kelei Cao$^{*}$,
Ross Maciejewski\thanks{e-mail:rmacieje@asu.edu},
Shixia Liu$^{*}$
}
\affiliation{%
  \scriptsize
   $^*$ School of Software, BNRist, Tsinghua University\\
   $^{\dag}$ Microsoft\\
   $^{\ddag}$ Bloomberg L.P.\\
   $^{\S}$ Computer Science, Arizona State University
}

\teaser{
\centering
 \vspace{-4.5mm}
\includegraphics[width=\linewidth]{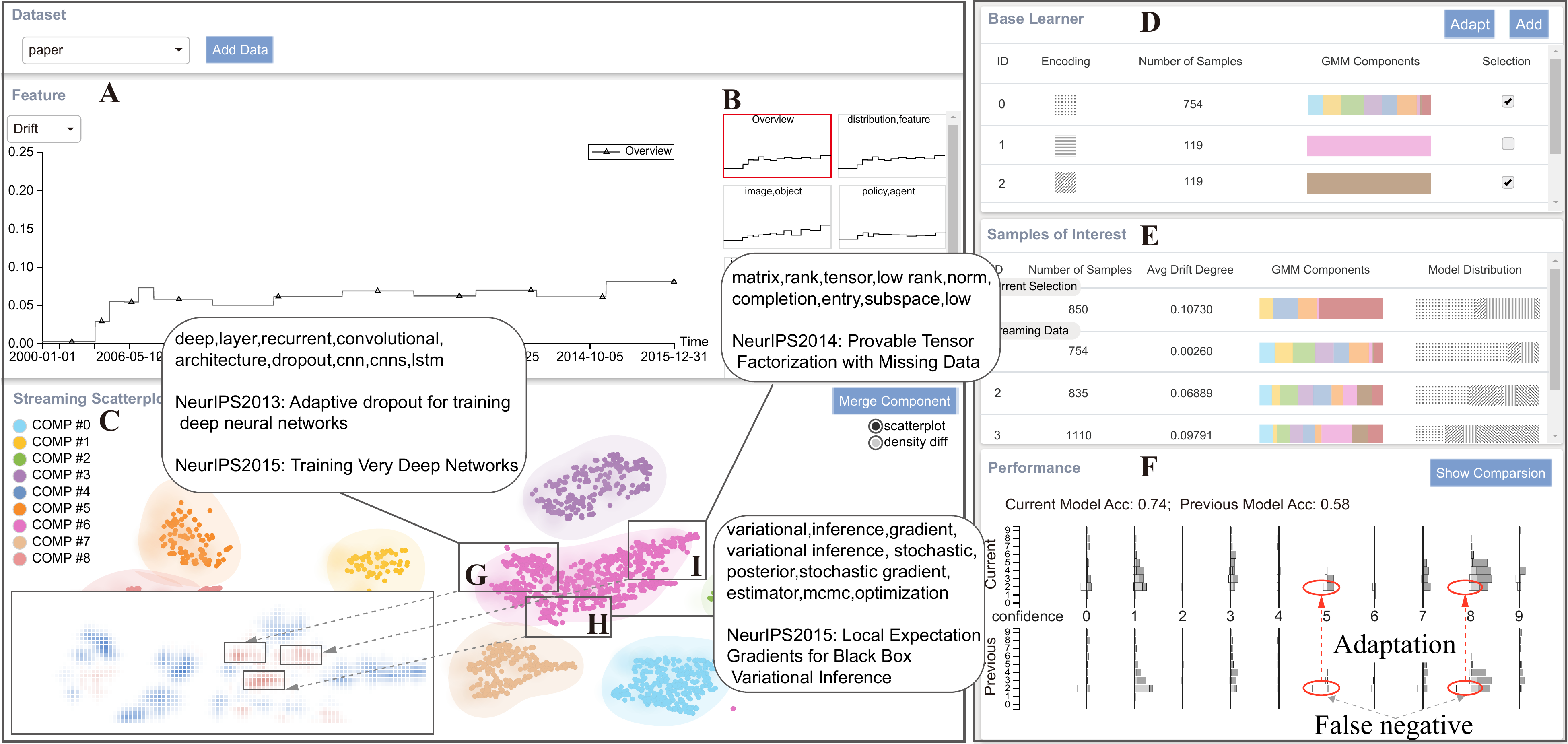}
{
\rmfamily
\put(-370,-10){(a) Stream-level visualization.}
\put(-140,-10){(b) Prediction-level visualization.}
}
\vspace{-1.5mm}
\caption{
\weikairevision{DriftVis: A visual analytics system for detecting, explaining, and correcting for concept drift: 
(a) The stream-level visualization consists of a line chart for drift degree (A), a feature selection list (B), and a streaming scatterplot (C) to visualize the drift and data distribution change over time (e.g., density increases in G, H, and I);
(b) The prediction-level visualization consists of a base learner view (D), a samples of interest view (E), and a performance view (F) to explore the impact of drift adaptation on the model's accuracy.}
}
\vspace{-2.5mm}
\label{fig:teaser}
}

\abstract{
Concept drift is a phenomenon in which the distribution of a data stream changes over time in unforeseen ways, causing prediction models built on historical data to become inaccurate.
While a variety of automated methods have been developed to identify when concept drift occurs, there is limited support for analysts who need to understand and correct their models when drift is detected. In this paper, we present a visual analytics method, DriftVis, to support model builders and analysts in the identification and correction of concept drift in streaming data.
DriftVis combines a distribution-based drift detection method with a streaming scatterplot to support the analysis of drift caused by the distribution changes of data streams and to explore the impact of these changes on the model's accuracy.
\weikairevision{A quantitative experiment and} two case studies on weather prediction and text classification have been conducted to demonstrate our proposed tool and illustrate how visual analytics can be used to support the detection, examination, and correction of concept drift.
}


\begin{document}




\fontsize{9}{9} 

\firstsection{Introduction}
\maketitle
To date, artificial intelligence technologies have made immense strides in developing machine learning models (e.g., classifiers) for real-world phenomena~\cite{sacha2019vis4ml}. 
Applications of these models are found in fraud detection~\cite{jiang2019recent}, medical diagnosis~\cite{jordan2015machine}, sales predictions~\cite{dovvzan2012solving}, and countless other domains. 
Often, models have a built-in assumption where the mapping function of input data used to predict an output value (e.g., prediction label) is assumed to be static.
However, as time passes, the mapping between the input data and the output value may change in unforeseen ways~\cite{Lu2018survey}, \liu{which could be caused by the change of the data and label.} 
In these cases, predictions made by a model trained on historical data may no longer be valid, and the model accuracy will begin to decrease over time. This phenomenon is typically referred to as \emph{concept drift} in machine learning.
As more and more machine learning applications move towards streaming data, the potential for model failure due to concept drift becomes further exacerbated. \looseness=-1


To prevent the degradation of prediction accuracy, many drift analysis methods have been proposed in the field of machine learning~\cite{iwashita2018overview, Lu2018survey}.
However, existing methods typically only provide a numerical drift value \mc{at each time point}. While such measures can identify when drift occurs, a single numeric cannot explain why the drift is occurring. 
Analysts need to have tools that can do more than simply alert them to when concept drift seems to be occurring. 
They need ways to understand how \liu{data} distributions have changed over time, which samples cause drift, and how the training samples can be adjusted for model building.
These underlying challenges of concept drift lend themselves well to a visual analytics paradigm in which the data and model can be \shixia{readily} evaluated for drift (``analyze first''). 
\weikairevision{When drift is detected, changes in data distribution can be highlighted (``show what’s important''), and an analyst only needs to label those drifted samples to update the model, which can save human effort.}
Once the analyst is satisfied, the system can return to the analysis state (``analyze again''). As such, we propose a visual analytics system, DriftVis, for detecting, explaining, and correcting for concept drift.



DriftVis has been designed to support the identification of concept drift, visually explain the drift analysis results, and enable the analysis of the root causes of such drift through a tight integration between the drift detection method and coordinated multiple views, creating a comprehensive visual analytics environment. 
\liu{As the change of data distribution is the root source of concept drift~\cite{Lu2018survey}, in this work, we focus on analyzing concept drift caused by distribution drift. Accordingly, 
the} drift detection method quantifies concept drift using a distance function that measures the difference between the distributions of the historical data used in model training and the incoming data.
In our implementation, the data distribution is modeled by an \mc{incremental} Gaussian mixture model (GMM). 
The energy distance, which is \mc{an} extension of the $L_2$ distance for measuring high-dimensional distributions, is utilized to compute the dissimilarity between two distributions~\cite{szekely2013energy}. 
This drift detection method is used to support stream- and prediction-level visualizations.
The stream-level visualization uses a drift degree line chart and a streaming scatterplot to disclose when and where concept drift might occur.
If drift is detected, analysts can utilize the streaming scatterplot to compare the distributions over time.
This scatterplot combines a constrained t-SNE and a multi-step animation to explain the distribution evolution.
The prediction-level visualization reveals the performance-related information of the prediction model.


In order to demonstrate the capabilities of our system, we have conducted two case studies exploring the drift in weather prediction and academic paper topic evolution.
A demo prototype of the system is available at \url{http://conceptdrift.thuvis.org/}.
The main contributions of our work include:
\begin{compactitem}
\item\noindent{\liu{\textbf{A visual analytics system} for detection, examination, and correction of concept drift.}}
\item\noindent{\liu{\textbf{A distribution-based drift detection method} to quantify the drift degree over time.}}
\item\noindent{\textbf{A streaming scatterplot} to explore the distribution changes over time and identify the root cause of concept drift.}
\end{compactitem}



\section{Related Work}
\label{sec:related-work}

%
%


\subsection{Concept Drift Analysis}
Research related to concept drift can be categorized into two main topics: drift detection and drift adaptation~\cite{Lu2018survey}.

\noindent \textbf{Drift Detection.} 
Error-rate-based methods and data-distribution-based methods are the most popular drift detection methods. 
The error-rate-based methods~\cite{baena2006early, gama2004learning} monitor the online error rate of a model based on ground-truth labels.
A drift is detected if there is a significant increase in the error rate.
These methods are less efficient in practice as it is often challenging to acquire high-quality data labels for calculating the error rate of streaming data.

Distribution-based techniques do not rely on data labels.
Instead, these techniques directly detect concept drift by calculating the distribution change \ross{with respect to} the \mc{streaming} data, which is a major cause of concept drift.
A key challenge for distribution-based techniques is how to measure the distribution change by calculating the distance between two data distributions.
Kifer et al.~\cite{kifer2004detecting} used total variation, which is one of the earliest attempts to use a distance function for drift detection.
Webb et al.~\cite{webb2016characterizing, webb2018analyzing} measured the distribution difference using the Hellinger distance and total variation.
\mc{In their work,} the drift degree was measured in different attribute subspaces, such as joint distribution and class distribution.

In our system, we utilize a distribution-based concept drift detection method, as these methods do not need ground-truth labels.
Furthermore, distribution-based methods can identify the samples in the streaming data that cause concept drift, which is useful for facilitating the understanding of concept drift.
Specifically, we model the streaming data distribution by an \mc{incremental} GMM and apply the energy distance \ross{function}~\cite{rizzo2016energy} to measure the distance between distributions.\looseness=-1






\noindent \textbf{Drift Adaptation.} 
Drift adaptation focuses on improving the model performance on \mc{the streaming data} by updating the existing learning model to \ross{account for} the drift. 
Existing methods of drift adaptation include model retraining and ensemble methods~\cite{Lu2018survey}.

Model retraining retrains a new model with a combination of the new data and old data to replace the old model. 
The main problem encountered with these types of methods is doc{determining} how much of the \mc{old} data will be discarded and how much of the new data will be utilized in training~\cite{bach2008paired}.
\yafeng{Ensemble methods comprise a set of base learners and} \mc{adjust the base learners} when adapting to the drift.
\mc{Typical adjustments include} adding new base learners, updating \mc{existing} base learners, removing unsatisfactory base learners, and \mc{changing} the weights of base learners in the ensemble model.
Hulten et al. proposed a tree-based method, CVFDT~\cite{hulten2001mining}, which replaced the sub-tree exhibiting poor performance with a retrained sub-tree.
The dynamically weighted majority method~\cite{kolter2007dynamic} down-weights the classifier that makes a wrong decision and up-weights the classifier that makes a correct decision.
When the ensemble \mc{model} makes a wrong decision, it will train a new base learner and add it to the ensemble, and the base learners with low weights will be removed.
Learn$^{++}.$Net~\cite{elwell2011incremental} re-weights the base learners based on the prediction performance on the latest streaming data to prevent adding new base learners too frequently.

\yafeng{Compared with model retraining, ensemble methods are more efficient because they do not require a full retraining of the model.
Furthermore, ensemble methods are capable of handling reoccurring drift by reusing old base learners.}
In DriftVis, we employ the method proposed by Kolter et al.~\cite{kolter2007dynamic} due to its competitive performance and flexibility.



%


\subsection{Visual Performance Analysis}
Existing efforts in visual performance analysis of machine learning models can be classified into two categories~\cite{choo2018visual,liu2017towards}: 1) understanding the model performance; 2) debugging and improving models. 

\noindent\textbf{Understanding model performance}. 
There are a variety of visualization tools that have been developed to support analysts in exploring the performance of machine learning models. 
Tzeng et al.~\cite{Tzeng2005} developed a visualization tool to support the performance analysis of artificial neural networks utilizing a graph-based visualization.
ModelTracker~\cite{amershi2015modeltracker} was developed to convey both the overall and instance-level performance of a binary classifier. 
Squares~\cite{ren2017squares} extended the work from ModelTracker and was designed to illustrate and compare the prediction score distributions of multiclass classifiers.
More recently, there has been a major focus on supporting the understanding of the performance of deep neural networks~\cite{choo2018visual,Hohman2019Visual,yuan2021survey}, such as convolutional neural networks~\cite{bilal2018convolutional,hohman2020summit,jia2019visualizing,rauber2017visualizing,wongsuphasawat2018visualizing}, recurrent neural networks~\cite{ming2017understanding,strobelt2018lstmvis}, deep Q-networks~\cite{wang2019dqnviz}, and deep generative models~\cite{kahng2019gan}.
While the majority of work has focused on supporting specific models, there has also been work focusing on developing model-agnostic interactive visualization tools to support model understanding~\cite{ming2019rulematrix,wexler2020if,zhang2019manifold}.
In general, the goal of many current systems is to support analysts in understanding the performance characteristics of models from multiple perspectives and illustrate which model features contribute to the performance.

\noindent\textbf{Debugging and improving models}.
Along with tools for understanding models, a variety of tools have been developed to support debugging and improving machine learning models.
These efforts focus on exploring the reasons why a learning process does not work as expected and supporting analysts in refining the model.
Liu et al.~\cite{liu2017towards} proposed a directed-acyclic-graph-based visualization method to illustrate how data flows through a network in the training process and thus facilitates the diagnosis of a failed training process.
DeepEyes~\cite{pezzotti2018deepeyes} was developed to help analysts identify the potential issues that deep neural networks may experience in the training process, such as redundant filters and inadequately captured information. 
Other research has explored model diagnosis of a variety of deep neural networks, such as deep generative models~\cite{liu2018analyzing,wang2018ganviz}, recurrent neural networks~\cite{kwon2019retainvis}, and deep sequence models~\cite{ming2020protosteer,strobelt2019seq}.

One current drawback with model debugging and performance improvement is the lack of generalizability. Models (or classes of models) often have specific parameters and features that are not completely generalizable.
As a result, most existing methods of visual model performance analysis are model-dependent. 
However, concept drift is a generalizable problem that focuses on understanding how data has shifted over time. Our system is designed around understanding shifts in streaming data and exploring distributions, making our proposed system model-independent.
This is similar to current work that focuses on analyzing the robustness of learning models in terms of data quality, such as understanding model vulnerabilities to adversarial examples~\cite{cao2020analyzing,liu2018robustness,ma2020explaining} and detecting out-of-distribution samples~\cite{Chen2020}.
These methods also compare training data with test data, identify outliers, and then analyze their effects on the model performance.
While inspired by the same motivation of enhancing the training dataset, our work differs from these methods in two ways. 
First, instead of considering only one static test dataset, we handle a stream of test data, where new test data keeps on coming in and then is appended to the existing data. 
Second, we measure the concept drift using a distribution-based method, which models the streaming data distribution by an incremental GMM. 
To illustrate the drift over time and the root cause of such drifts, a streaming scatterplot is developed, which combines the advantages of the constrained t-SNE and a density map. 

\begin{figure*}[!tb]
\centering
{\includegraphics[width=\linewidth]{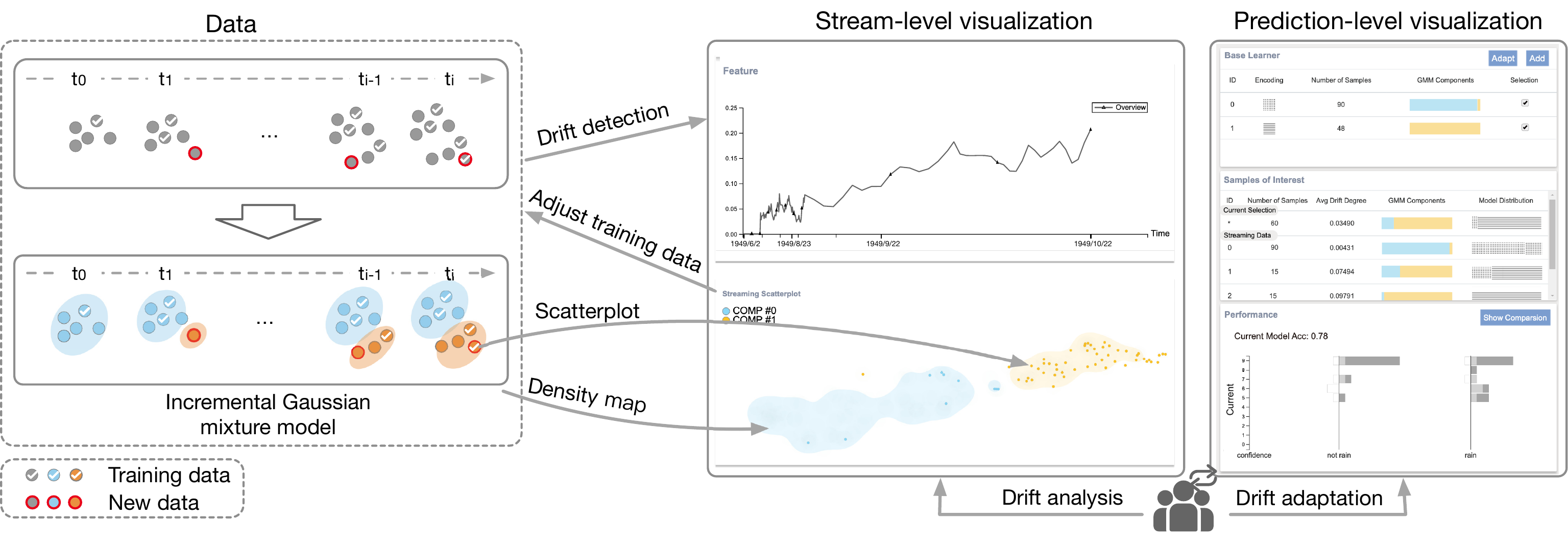}}\hfill
\vspace{-3mm}
\caption{DriftVis overview.
When data streams in, an incremental Gaussian mixture model is first employed to model the data, and the output is further used in drift detection and data visualization.
The stream-level visualization displays the drift degree and the time the distribution changes to support drift analysis.
The prediction-level visualization facilitates the understanding of the model performance for drift adaptation.
\yafeng{The training data in the adapted prediction model will be used for drift detection in the next iteration.}
}
\vspace{-5mm}
\label{fig:system-overview}
\end{figure*}

\section{Requirement Analysis}



This work is driven by previous research on concept drift and our discussions with a researcher from the China Meteorological Administration (E$_1$) and three machine learning practitioners (E$_2$, E$_3$, and E$_4$) regarding their needs for model deployment in real-world applications.
E$_1$ works on weather predictions.
E$_2$ focuses on predicting whether a computer motherboard meets the quality standards for a computer manufacturing company, Inventec.
E$_3$ and E$_4$ are machine learning researchers and are interested in classifying conference papers to facilitate their research,
and E$_4$ is a co-author of this paper.
A central problem each of these experts faces after deploying a machine learning model is how to monitor the model performance over time and decide when the model needs to be updated if an unsatisfactory performance is identified.

To understand the experts' needs, we engaged in weekly free-form discussions with our \yafeng{experts} for two months to learn about the systems they were currently using for model performance analysis, and what challenges they faced when using these systems.
In our discussions, the \yafeng{experts} commented that there are already several cloud-based model performance monitoring services.
Two widely-known services are Google's continuous evaluation service~\cite{Google_continue} and the data drift detection service on Microsoft Azure~\cite{MS_drift}.
These services can alert analysts when drift occurs.
However, they do not support the exploration of where the data distribution has changed.
This requires the analyst to spend more time investigating
\yafeng{which subset of the most recent data could benefit the prediction model the most}
and then labeling the data before retraining the model. 
Our experts noted that if the system could help identify where the distribution changes occurred, the amount of labeling could be reduced. 
\mc{This is favored by the experts because the cost of acquiring a large number of data labels is usually high.}



Based on the literature review, the observed limitations of current drift detection methods, and the discussions with experts, we summarized the requirements for the visual analytics system as follows:



\noindent \liu{\textbf{R1. Detecting when concept drift happens}.}
Concept drift is one of the major causes of model accuracy degradation over time.
To tackle this issue, many methods have been proposed to track and detect when drift occurs in streaming data~\cite{gama2014survey,Lu2018survey,street2001streaming}.
All the experts expressed that they required a convenient way to alert them to the potential occurrences of concept drift during the running process of their deployed machine learning models. 
For example, E$_2$ once found that his quality control algorithm wrongly judged some qualified motherboards to be unqualified because a new batch of motherboards used in the assembly line did not have the same board height as before.
Thus, it is critical to detect potential concept drifts in \mc{streaming data} and inform the analyst to minimize the decrease in accuracy and quickly recover from the drift process.
\noindent \textbf{R2. Analyzing where and why drift happens.}
All the experts expressed a strong need for analyzing where and why drift occurs after it is detected and understanding how the detected drift will influence the model performance. 
Thus, a simple numerical measure of drift degree is not sufficient.
For example, experts usually want to know how each feature has changed over time.
E$_3$ commented, ``We know which features are important to the model, and if there are significant changes in these features, the drift is likely to affect the model performance.''
Other useful information to explore \mc{includes} the distribution of the data and their changes over time.
The experts wanted to quickly locate the regions dominated by new data or the regions where the data distributions have changed. 
In these cases, the model performance is likely to have a drop.
In addition to exploring where drift happens (e.g., which feature and where it is in the data distribution), the analysis of why \mc{the drift} happens is also important. 
E$_4$ said, ``Knowing why the drift occurs really helps to take necessary actions - sometimes the action \yafeng{may} not even be required on the model side. 
For example, several months ago, I found that the model performance degradation was simply caused by the changed position of the camera. 
It was moved to another angle, resulting in the incoming video not being the same as the training dataset.''
The need to analyze where and why is \yafeng{also} well aligned with previous research on concept drift analysis~\cite{liu2017regional,Lu2018survey, wang2020conceptexplorer}.

\noindent \textbf{R3. Overcoming drift and improving performance.}
Having an understanding of the drift is not the end of the analysis.
Experts needed to appropriately adapt the model to the new data and improve the performance, which is called drift adaptation.
\yafeng{The most straightforward adaptation approach is to retrain the model with additional labeled streaming data.}
However, the \yafeng{experts} were not in favor of this method.
E$_3$ commented, ``For us, a lighter-weight method is more practical. It is inefficient to retrain the whole model as it often takes hundreds of GPU hours.''
E$_2$ also desired to shorten the time of drift adaptation because 
the quality control of the computer motherboard assembly line is crucial and would have a large influence on the brand.
\yafeng{Fast reactions to any detected issues on the computer motherboard assembly line can reduce company losses.}
E$_1$ further pointed out that there were sometimes reoccurring patterns in the weather data, such as revolving seasons and day-night alternations.
Thus, it is desirable to have a drift adaptation method to reuse part of the old model.
This is also reflected in the current research trend that ``adaptive models and ensemble techniques have played an increasingly important role in recent concept drift adaptation developments~\cite{Lu2018survey}.''\looseness=-1


\section{Design of DriftVis}
\label{sec:DriftVisDesign}
\label{sec:system}

DriftVis is designed to detect, analyze, and overcome concept drift.
It consists of three modules: drift detection, drift visualization, and drift adaptation (Fig.~\ref{fig:system-overview}).

The \textbf{drift detection} module compares the training dataset with the \yafeng{latest} streaming data and calculates the drift degree over time (\textbf{R1}).
An incremental GMM is employed to model the data distribution, and a distribution-based drift detection method is developed to calculate the drift degree.
Detecting drift in streaming data serves as the basis for the subsequent analysis, including discovering the root cause of the detected drift and adapting the model to compensate for it.\looseness = -1

The \textbf{drift visualization} module provides an exploratory analysis of when, where, and why drift happens (\textbf{R1}, \textbf{R2}).
Such analysis is primarily supported by a streaming scatterplot that shows the data distribution over time.
Combined with the drift degree line chart and a feature selection list, analysts can make informed decisions on how to adapt the model to overcome the drift.


The \textbf{drift adaptation} module adapts the learning model to \mc{the unforeseen change in streaming} data (\textbf{R3}).
In \minorchange{DriftVis}, we employ the ensemble method developed by Kolter et al.~\cite{kolter2007dynamic} due to its competitive performance and flexibility.
In particular, we train a set of base learners on different subsets of the historical data.
\mc{These base learners are combined with different weights to adapt to the new data} and improve performance. 

Our \yafeng{contributions} mainly focus on the first two modules, so we introduce them in detail as follows. 

\subsection{Drift Detection}
\label{sec:DriftDetectionMethod}


One major source of concept drift is the unexpected changes in the intrinsic distribution of the data stream~\cite{iwashita2018overview, Lu2018survey}.
Thus, a key challenge to compute the drift degree is how to measure the difference between data distributions. 
\textbf{Energy distance} is motivated by the potential energy between objects in a gravitational space~\cite{rizzo2016energy}.
The distance is zero if and only if the two distributions are identical.
The major advantage of energy distance is that its calculation is linear to the number of dimensions and can be easily scaled to high dimensions~\cite{goldenberg2018survey}. 
As a result, we employ energy distance to measure the drift degree.
For two groups of data samples $X = \{{x}_1,\ldots,{x}_n\}$ and $Y = \{{y}_1,\ldots,{y}_m\}$, the energy distance of their underlying distributions is defined as:
\begin{equation}
\label{eq:drift_origin}
d(X,Y)=(2A-B-C)/2A,
\end{equation}
where $A=\frac{1}{mn}\sum_{i=1}^{n}\sum_{j=1}^{m}\lVert {x}_i-{y}_j\rVert$ is the average of the pairwise distance between two groups of samples, $B=\frac{1}{n^2}\sum_{i=1}^{n}\sum_{j=1}^{n}\lVert {x}_i-{x}_j\rVert$ and $C=\frac{1}{m^2}\sum_{i=1}^{m}\sum_{j=1}^{m}\lVert {y}_i-{y}_j\rVert$ are the averages of pairwise distance within $X$ and $Y$, respectively.
Here $\lVert{x}\rVert$ is the L2-norm of vector ${x}$\looseness=-1.
%

In DriftVis, the drift degree at time $t$ is \yafeng{measured} by the energy distance between the training set $\hat X$ and the streaming data $X_t$ at time $t$.
\mc{In practice, data samples may come one at a time, such as daily weather data.
In that case, $X_t$ is set as the streaming data in a sliding window ending at time $t$ to reduce noise and get a more accurate drift degree.}
In the development of DriftVis, we found that directly calculating the drift degree between $\hat X$ and $X_t$ may overestimate the drift.
This typically occurs when $\hat X$ contains several clusters, while $X_t$ only contains a subset of these clusters.
For example, $\hat X$ may contain the weather \yafeng{records} for a whole year, \yafeng{but} $X_t$ only contains one month of data.
In this example, the calculated distance is large because $X_t$ lacks the samples from other months.
However, the model performance drop is not large since the distribution of $\hat X$ already covers that of $X_t$. 
As a consequence, directly calculating the distribution distance can potentially generate false alarms.

To tackle the issue of drift over-estimation, it is desirable to compare $X_t$ with only the training samples that are similar to them. 
The key challenge is, for each newly arrived data sample, how to select the similar training samples, and how many samples are enough. 
A straightforward solution is to use the $k$-NN algorithm to select the top-$k$ similar training samples.
Although this method is able to reduce the drift over-estimation to some extent, experts \yafeng{have to tune}
the parameter $k$ for different datasets and even for different samples to achieve good results, which is impractical in applications.
To solve this problem, we let the data speak for itself, i.e., automatically determining the similar samples based on the data sample clusters in the dataset.
\mc{Compared with $k$-NN, the number of clusters of a clustering algorithm can often be automatically determined with unsupervised model selection methods, such as Akaike information criterion (AIC)~\cite{akaike1998information} and Bayesian Information Criterion (BIC)~\cite{schwarz1978estimating}.}
The intuition behind this solution is that in the same cluster of the dataset, the samples are similar.
Thus, at each time $t$, we first incrementally cluster the data samples $X_t$ at time $t$ based on the clustering result before $t$.
Then, based on the new clustering result, for each sample $x_t^i$ in $X_t$, we select the training samples that are in the same cluster of $x_t^i$ as its similar samples.

\mc{Due to streaming nature of the data, we employ an incremental clustering method in DriftVis, i.e., the incremental GMM algorithm~\cite{engel2010incremental}}.
A GMM-based clustering method is utilized because GMM is able to approximate almost any continuous distribution where the value of each feature is continuous~\cite{bishop2006pattern}.
The incremental GMM algorithm can be split into an offline clustering step for the training dataset $\hat X$ and an online clustering step that ingests data samples $X_t$ for each time point $t$:

\begin{figure*}[!tb]
\centering
{\includegraphics[width=\linewidth]{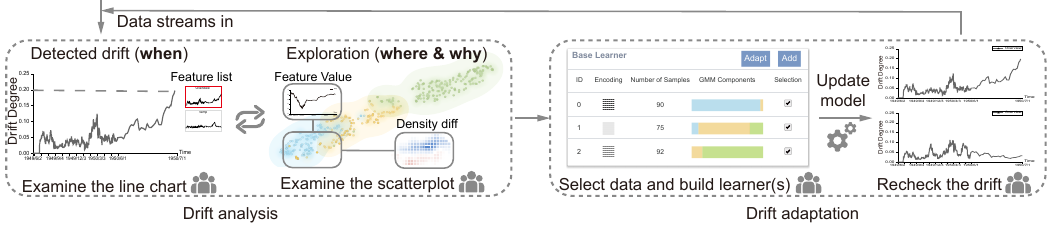}}\hfill
\vspace{-3mm}
\caption{DriftVis analysis workflow. When drift is detected, analysts can explore when, 
where, and why it happens with the help of stream-level visualization.
To adapt to the drift, analysts can select data samples to build learner(s) and then update the model.}
\label{fig:workflow}
\vspace{-4mm}
\end{figure*}

\noindent \textbf{Offline Clustering.}
A GMM is built to cluster the data samples in $\hat X$, where each cluster is a Gaussian component $N(\cdot \mid \mu, \Sigma)$ with mean $\mu$ and covariance $\Sigma$.
The number of clusters is determined by the Bayesian Information Criterion (BIC)~\cite{schwarz1978estimating}.

\noindent \textbf{Online Clustering.} 
The major purpose of this step is to incrementally cluster the data samples in $X_t$. 
For each data sample $x_t^i$ in $X_t$, we need to decide 
\yafeng{which existing cluster it belongs to or if a new cluster should be created when it is not close to any existing cluster.}
If the probability of this sample belonging to an existing Gaussian component $p=N(x_t^i|\mu, \Sigma)$ is higher than a threshold (0.95 in DriftVis), we assign the data sample to the cluster with the maximum probability and update the mean and the covariance of the corresponding Gaussian component~\cite{engel2010incremental}.
Otherwise, there should \weikai{exist}{be} a new Gaussian component \weikai{which}{that} better fits the current data sample.
Constructing a Gaussian component from a small number of samples cannot accurately estimate the component's mean and covariance, and will bring instability to the clustering result.
To solve this problem, we temporarily assign these samples to the Gaussian components with the maximum probability.
\weikairevision{When enough such samples are identified, we build a new GMM using these samples and then merge the new components into the current clustering results. In DriftVis, to get enough samples and to avoid creating noisy components, we set the threshold to be half of the average component size or we allow the analyst to set a specific threshold value.}
Based on the clustering result at $t$, we calculate the drift degree $D_t$ between $X_t$ and $\hat X$ as a weighted sum of the drift degree for each cluster:\looseness=-1
\begin{equation}
\label{eq:drift_weighted}
D_t = \sum_i \frac{|X_t^i|}{|X_t|}d(X_t^i, \hat X^i),
\end{equation} 
where $|\cdot|$ is the number of samples in a set and $d(\cdot)$ is the energy distance.
$\hat X^i$ and $X_t^i$ are the samples of the $i$-th cluster in $\hat X$ and $X_t$, respectively.
For a cluster that satisfies $|\hat X^i|=0$, we set the drift $d(X_t^i, \hat X^i)=1$ because the data samples in $X_t^i$ are unlike other samples in the training dataset and will probably cause model performance drop.
In this case, there should be a larger drift degree.

\subsection{Drift Visualization}
\label{sec:visualization}



To support the analysis of \mc{when, where, and why drift happens (\textbf{R1}, \textbf{R2}),} DriftVis employs two visualization components, a stream-level visualization (Fig.~\ref{fig:teaser}(a)) and a prediction-level visualization (Fig.~\ref{fig:teaser}(b)).
The stream-level visualization discloses the distribution changes of the data stream over time. It 
supports the exploration purely of the data, including the distribution of data samples, the drift detection results, and some of the raw data contents (e.g., the raw text before feature extraction).
\yafeng{The prediction-level visualization focuses on the performance-related information of the prediction model and reveals how the adaptation to the drift affects the model performance (\textbf{R3}).}
The drift detection and adaptation methods support the two visualization components and serve as the foundation for the concept drift analytics environment.\looseness=-1 


Fig.~\ref{fig:workflow} illustrates the typical analysis workflow as data streams in. 
The workflow begins with an analyst checking the drift degree line chart \yafeng{to detect any obvious drift increase (``when'')}.
\yafeng{In cooperation with the ``when'' analysis, the analyst can further analyze ``where'' and ``why'' the drift happens on the streaming scatterplot, which illustrates}
if the \yafeng{incoming data} forms new components, \mc{deviates from the existing components or causes density increase/decrease in some regions.}
After \yafeng{confirming a drift}, the analyst can select data samples on the scatterplot to build new base learners. 
Using the available base learners, the analyst updates the ensemble and 
checks if the model performance has improved using the \yafeng{prediction-level} views. 
This workflow can be applied iteratively as new data streams in.\looseness=-1

\subsubsection{Stream-Level Visualization}\label{subsec:streamLevelVis}
The stream-level visualization is designed to illustrate and explain concept drift \yafeng{- specifically, the calculated drift degree and the density change}. It consists of a drift degree line chart (Fig.~\ref{fig:teaser}A), a feature selection list (Fig.~\ref{fig:teaser}B), and a streaming scatterplot (Fig.~\ref{fig:teaser}C).
The drift degree line chart is used to alert the analyst of possible occurrences of concept drift by displaying the calculated drift degree over time (\textbf{R1}). 
The line chart coordinates with the feature selection list to turn on and off the drift degree lines calculated on different features.
When any single feature is selected, the y-axis can switch between the drift degree and the feature value.
\weikairevision{We used the line chart because it is the most common form of time series visualization~\cite{Heer2009, liu2017towards}.\looseness=-1}

As described in requirement \textbf{R2}, it is critical to know where \yafeng{and why} drift occurs by exploring the change of data distribution over time.
We use a GMM-based constrained t-SNE to \mc{continuously} project the streaming data onto a 2D space and have designed two visualization modes, scatterplot and density diff, to present the data for distribution-based analysis. 

%
\noindent\textbf{GMM-based constrained t-SNE.} 
The GMM-based constrained t-SNE is designed to 1) preserve the \yafeng{sample} similarities in the original high dimensional space, 2) explain the distribution of each Gaussian component, and 3) minimize the movements of previous data points in the t-SNE plot when new data \yafeng{streams in}. 
The state-of-the-art solution to achieve the aforementioned goals is the supervised t-SNE developed by Choo et al.~\cite{choo2013utopian}, where \yafeng{samples} in the same class are grouped together by reducing their distances \yafeng{using} a shrink factor.
Although shrinking the projection within each group can reduce the visual clutter between groups and \yafeng{improve} readability, \yafeng{it} does not consider the dispersion of each group and 
\yafeng{could also not maintain stability when the projection updates with additional data.}
To overcome these limitations and well convey the distribution patterns of Gaussian components, we modify the shrink factor with respect to the dispersion of each component \yafeng{and add two constraints to improve the stability of the shape of Gaussian components and data positions.}
\looseness =-1

\mc{As we assume that the ground-truth labels are unavailable when calculating the projection,} the shrink factor uses the component label inferred from the incremental GMM. 
Its value \yafeng{is weighted by} the component dispersion that highly dispersed components will shrink more.
In DriftVis, we use entropy to measure dispersion.
For each Gaussian component $C_k$ whose covariance matrix is $\Sigma_k$, the shrink factor $\alpha_k$ is determined \yafeng{as follows:}
\begin{equation}
    \alpha_k = \alpha\cdot\left(1- \beta\cdot \frac{H[C_k]}{\max_{k'}\{H[C_{k'}]\}}\right).
\end{equation}
Here \mc{$\alpha$ is the maximum allowed shrinking} factor, $\beta$ is a \mc{weighting} factor, and $H[C_k]$ is the entropy of the Gaussian distribution of component $C_k$.
In practice, \mc{to ensure that the calculated shrink factor does not exceed the maximum, $\alpha$}, we enforce $H[C_k]$ to be no smaller than a small positive $\epsilon$. Thus, $H[C_k] =\max( \frac{1}{2}\ln{|\Sigma_k|} + \frac{D}{2}(1 + \ln{2\pi}), \epsilon)$, and $D$ is the size of the feature dimension. 
For the most dispersed component, $\alpha_k = \alpha\cdot(1-\beta)$, and 
\yafeng{for the least dispersed component $\alpha_k \approx \alpha$.}
Adding this factor to the distance calculation yields, 
\begin{equation}\label{eq:dist}
\text{dist}({x}_i, {x}_j) =\left \{
\begin{array}{ll}
\alpha_k\cdot\|{x}_i- {x}_j\| & {x}_i, {x}_j \in C_k \\
\|{x}_i- {x}_j\| & \text{otherwise}
\end{array}.
\right.
\end{equation}

To further improve the stability of the projection results, we add two constraints into the supervised t-SNE:

The first is a shape constraint that maintains the shape of each Gaussian component to be stable between adjacent time points.
Suppose we have $n$ \yafeng{samples} of original data and $m$ new \yafeng{samples} from the streaming data.
The $n$ original \yafeng{samples} form $k$ original Gaussian components.
As the shape is hard to measure, we approximately maintain the relative distances between the original $n$ data samples and the component centers. 
This is achieved by minimizing the KL-divergence between $P_c$ and $Q_c$ $\in \mathbb{R}^{n\times k}$, which represent the joint probability distributions of the similarities between the $n$ original samples and the $k$ center constraint points in the high dimension and low dimension, respectively.
Each center constraint point is taken as the gravity center of the Gaussian distribution in the high dimensional space and its corresponding low dimensional projection in the previous iteration.
Following t-SNE, the distributions $P_c$ and $Q_c$ are defined as:
\begin{equation}
    \begin{array}{rl}
       (P_c)_{ik} &= w_k\cdot\exp(-\text{dist}({x}_i, {x}_k)/2\sigma^2)\\
       (Q_c)_{ik} &= (1+ \|{y}_i- {y}_k\|^2)^{-1} 
    \end{array}.
\end{equation}
Here, $w_k$ is the weight for the $k^{th}$ center constraint and is set to $\sqrt{\lvert C_k\rvert}$, where $\lvert C_k\rvert$ is the sample number in the $k^{th}$ Gaussian component. ${x}_i$ is the high dimensional vector of the original instance, and ${x}_k$ is the high dimensional center of the Gaussian component $C_k$,
and ${y}_i$ and ${y}_k$ are their projected low dimensional coordinates. $\text{dist}({x}_i, {x}_k)$ uses Eq.~(\ref{eq:dist}).

The second constraint is to maintain the positions of the original samples.
\lu{To improve the run-time efficiency, when sample size $n$ is greater than 500, we use blue noise sampling~\cite{balzer2009capacity} to select $n^\prime$ (500) samples from $n$ original samples and introduce a virtual constraint point for each.}
$P_s$, $Q_s$ $\in \mathbb{R}^{n\times n^\prime}$ represent the joint probability distributions of the similarities between the $n$ original samples and the $n^\prime$ selected samples.
When sample $x_i$ is used as the $k^{th}$ virtual constraint point, $(P_s)_{ik} = 1$, and this will restrict the change of the projected coordinates of the $n^\prime$ selected samples and maintain the position stability of the overall projection.

Combining the above two constraints and the supervised t-SNE, we obtain the projection by minimizing the objective function:
\begin{equation}\label{eq:cost}
    \lambda\cdot KL(P\|Q) + \varphi\cdot KL(P_c\|Q_c) + (1 - \lambda - \varphi)\cdot KL(P_s\|Q_s).
\end{equation}
Here $KL(\cdot\|\cdot)$ is the KL-divergence between two distributions.
The first term \mc{is the original optimization goal of supervised t-SNE that minimizes the difference between the high-dimensional distribution $P$ and the 2-D distribution $Q$ of all the $m+n$ \yafeng{samples}.
The second and third terms are the shape and the position constraints, respectively.}

\noindent\textbf{Scatterplot and density diff.}
To visually explain the distribution and the drift, the streaming scatterplot utilizes two modes, scatterplot and density diff.
\emph{Scatterplot mode} uses a density representation to display the historical data distribution and scattered points to represent streaming data. 
All data samples are projected using the GMM-based constraint t-SNE projection but rendered differently.
Historical data is rendered using a density map where a lighter color represents a lower density. 
On top of the density map, the streaming data is displayed as scattered points. 
Categorical colors are used to separate the Gaussian components.
Fig.~\ref{fig:teaser}C shows an example of the scatterplot mode 
for analyzing
\mc{the papers from NeurIPS.}

To illustrate the change of data distribution, we utilize a two-stage animation
to \yafeng{exhibit} the appearance of incoming data samples.
The first stage updates the projection of historical data and the density map.
Historical data points move from their original positions to their updated ones as the new streaming data is used in the projection calculation.
As this movement occurs, a new density map will be constructed using the updated projection of the historical data points.
The second stage draws the new streaming data where data points are grouped based on their Gaussian component assignment and displayed by group.

The \emph{density diff mode} is used to illustrate \mc{the density change} of data distribution by visualizing the density difference between two \yafeng{datasets, which can be} consecutive data batches 
\yafeng{or two batches selected by the analyst.}
To gauge the density difference, this view uses grids to represent the 2D space and calculates the percentage of data points falling into each grid in the \yafeng{latest streaming data.}
For the grid on the $i^{th}$ row and $j^{th}$ column with $n$ data points, its \yafeng{normalized} \mc{density} is
$grid^t_{i,j} = n /N$ where $N$ is the number of data points at time $t$.
To produce a more legible visualization, the grid \yafeng{normalized density} is then \mc{smoothed by a halo effect~\cite{oelke2011visual} where each grid is extended into a 5x5 grids, and we assign 30$\%$ density to the peripheral halo area}.
A positive value means the grid has a higher density in the \yafeng{latest streaming data} and is colored red.
A negative value means the grid has a lower density in the \yafeng{latest streaming data} and is colored in blue.
The darker the color, the higher the density difference. 
Fig~\ref{fig:first_year} shows an example of the density diff view.

\weikairevision{\noindent\textbf{Justification.} 
We considered using either a flow map or a sequence of screenshots to show the distribution change over time. However, we found that the flow map of the data distribution cannot support a detailed distribution analysis at a particular timestamp, and the screenshot sequence makes it hard to identify the 
\lu{distribution change between}
screenshots~\cite{goldsberry2009issues}. Therefore, we adopted the streaming scatterplot because it can reveal the detailed distribution and support comparison~\cite{liu2016visualizing}. 
Analysts can switch between the \emph{scatterplot mode} and \emph{density diff mode}. We adopted this switching design because superposition can cause visual clutter, and juxtaposition would halve the view size and make it harder to investigate the distribution.
}

\subsubsection{Prediction-Level Visualization}
The prediction-level visualization consists of three views, a base learner view (Fig.~\ref{fig:teaser}D), a samples of interest view (Fig.~\ref{fig:teaser}E), and a performance view (Fig.~\ref{fig:teaser}F).

The \emph{base learner view} lists all available base learners that can be used in the ensemble model.
\yafeng{For each base learner, it shows the size of the training set and the proportion of the data points in the training set that belong to each Gaussian component. 
This proportion is represented by a colored bar using the same color scheme in the streaming scatterplot.}
Analysts can select a data subset to create a new base learner and choose which base learners to use in the ensemble model.\looseness=-1

\mc{The \textit{samples of interest view} is used to mark a subset of data as samples of interest, which can be revisited afterward.}
Similar to the base learner view, the colored bars show the percentage of Gaussian component labels in every batch.
To see the effect of each base learner on the samples, we calculate a model distribution and visualize it using the combination of each base learner's glyph pattern. 
The model distribution is a summary of the importance of the base learners in making predictions for this dataset. 
\yafeng{The importance of each base learner is gauged by the sum of the weights every training sample takes from the base learner.}
From the model distribution, the analyst can determine which base learner is important.
\weikairevision{We chose the table form here because it is simple and easy to understand~\cite{Muzammil2011DataAI}.}

The \mc{\textit{performance view} visualizes the model performance before and after adaptation to verify the effectiveness of drift adaptation.}
\yafeng{This view would only present on the data with labels.}
To enable the exploration of prediction accuracy and visualize the direct effect of the adaptation, we utilize the design from Squares~\cite{ren2017squares} in our performance view, as shown in Fig.~\ref{fig:teaser}F.
\mcr{Squares is able to provide a compact and effective comparison between classifiers before and after adaptation.}
It shows each class in a column, which contains a vertical axis annotated by its class label. 
The left-most vertical axis indicates the prediction confidence ranges.
Summary statistics (true positive, false positive, and false negative) for each class are encoded in stacked bars along the corresponding axis. 
The length of the bar represents the percentage of data samples predicted with the corresponding confidence score.
Dark grey indicates true positive, light grey indicates false positive, and no-fill bars on the left of the axis indicates false negative. 
To see if the prediction performance has improved, the performance view can change into comparison mode to show the performance statistics of the latest model and the model prior to this adaptation.

\subsubsection{Coordinated Interactions}
DriftVis provides coordinated interactions among different views for drift analysis. 
When the analyst hovers over a data sample or selects a group of samples, the line chart will highlight the time (on the x-axis) when these samples occur.
The analyst can also brush the line chart to see the brushed samples highlighted on the scatterplot. 
The streaming scatterplot itself also possesses a rich set of interactions for data exploration.
In order to investigate data details, such as the feature values and raw data, the streaming scatterplot uses pop-up windows to display the information of hovered or selected data samples. 
Lasso selection is supported to display a summary of the selected samples.
The streaming scatterplot also allows the analyst to hide Gaussian components when it is too crowded.

To decide which base learner to include in the ensemble, it is essential to know what data is used to train the base learner.
Hovering over a base learner highlights its training samples on the scatterplot, and if the analyst clicks on the base learner, these samples will be shown as selectable points so that they can be reused to train another base learner or be compared exclusively to the \yafeng{latest streaming data}.


Sometimes, an undesirable incremental GMM result may cause a false alarm on drift detection because some data are not compared with the closest distribution;
therefore, the drift degree is overestimated.
To reduce such false alarms, our system supports analysts to modify the incremental GMM result. 
Analysts can select some data in the scatterplot and click ``Merge Component'' to confirm that these points should belong to \mc{one component. The belonging relationship of other data samples will be incrementally updated.}

\section{Evaluation}
\label{sec:application}

\begin{figure*}[!t]
  \centering
{\includegraphics[width=\linewidth]{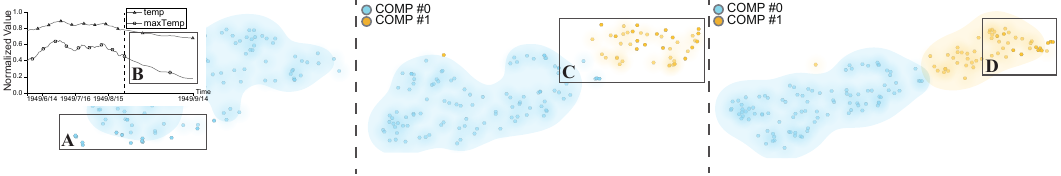}}
{\rmfamily
\put(-465,2){(a) Sept. 1 to Sept. 15.}
\put(-287,2){(b) Oct. 1 to Oct. 23.}
\put(-111,2){(c) Oct. 24 to Nov. 2.}
}
  \vspace{-5.2mm}    
  \caption{\weikairevision{The streaming scatterplot shows the density change from summer to early autumn in 1949:
  (a) the new data (A) came with a lower temperature (B);
  (b) a new component (C) was detected;
  (c) the following data (D) came and deviated from the yellow component.}}
  \label{fig:autumn_data}
   \vspace{-5.2mm}
\end{figure*}


\weikairevision{In this section, we conducted a quantitative experiment to evaluate the effectiveness of our drift detection method and then presented two case studies to demonstrate how DriftVis supports the identification, analysis, and correction of concept drift.}

\subsection{Quantitative Evaluation on Drift Detection}

\noindent \textbf{Experimental setting.} \mcr{We \mcr{employed} the widely-used experimental setting in drift detection methods~\cite{dasu2006information,FengGu2016}.
Two synthetic datasets, $D_1$ and $D_2$, were used.
Each dataset consists of 5 million data points, which follow $2$-D Gaussian distributions.
There are $99$ simulated concept drifts in each dataset.
The drifts were simulated by changing the mean and variance of the Gaussian distributions in $D_1$ and $D_2$, respectively.
We compared our method with three baseline methods: ITA~\cite{dasu2006information}, LDD-DIS~\cite{liu2017regional}, and TVD~\cite{webb2018analyzing}.
In our experiment, we used a grid search to obtain the best hyper-parameters.}

\noindent \textbf{Criteria and results.} \mcr{
We classified the detected drifts into four categories by comparing them with the ground-truth.
When a drift happens \lu{at} $t$ and is reported \lu{at} $t+\Delta t$, the drift is considered to have been successfully \textit{detected} if $\Delta t<w$, \textit{late} if $\Delta t\ge w$ but is reported before next drift happens, and \textit{missed} if it is not reported before \lu{the} next drift happens. Here, $w$ is the window size of the respective method.
\lu{A} \textit{false} alarm happens if 
\lu{a} drift is reported again before the next drift occurs.
In practice, we want to maximize the number of \textit{detected} drifts and minimize the other three categories.
Table~\ref{tab:result} shows the number of drifts in each category for each method.
To reduce randomness, we reported the numbers averaged over 10 runs.
Compared with the baselines, our method has detected the largest number of \textit{detected} drifts in both datasets, demonstrating an improvement over existing drift detection methods.
However, we noticed that, compared with $D_1$, our method performed worse in $D_2$.
The potential reason is that the energy distance mainly focuses on the mean ($D_1$) of the distribution instead of the variance ($D_2$).\looseness=-1
}

\begin{table}[!t]
\vspace{-0.5mm}
\caption{Comparison of drift detection results.} 
\vspace{-1mm}
\begin{tabular}{clcccc} 
\toprule 
Dataset & Method & Detected & Late & Missed & False \\ 
\midrule 
\multirow{4}*{$D_1$} & ITA~\cite{dasu2006information} & 75.5 & 12.3 & 11.2 & 4.5 \\
~ & LDD-DIS~\cite{liu2017regional}& 56.5 & 19.3 & 23.2 & 3.9 \\
~ & TVD~\cite{webb2018analyzing} & 75.5 & 19.9 & 3.6 & 7.0 \\
~ & Ours & \textbf{97.9} & \textbf{0.9} & \textbf{0.2} & \textbf{0.8} \\
\hline
\multirow{4}*{$D_2$} & ITA~\cite{dasu2006information} & 75.7 & 18.0 & \textbf{5.3} & 7.6 \\
~ & LDD-DIS~\cite{liu2017regional} & 44.0 & 22.2 & 32.8 & 12.2 \\
~ & TVD~\cite{webb2018analyzing} & 71.2 & 22.3 & 5.5 & \textbf{6.0} \\
~ & Ours & \textbf{77.5} & \textbf{14.9} & 6.6 & 6.4 \\
\bottomrule
\end{tabular}
\label{tab:result}
\vspace{-5mm}
\end{table}

\subsection{Case Study}

\subsubsection{Tabular Data: Weather Prediction}

\noindent\textbf{Dataset}.
This dataset is compiled by the U.S. National Oceanic and Atmospheric Administration (NOAA).
We use a preprocessed subset that has been used in previous concept drift detection research~\cite{ditzler2012incremental,elwell2011incremental}.
It contains 18,159 daily weather records from the Offutt Air Force Base from 1949 to 1999.
Each weather record contains eight measurements (``Temperature,'' ``Dew Point,'' ``Sea Level Pressure,'' ``Visibility,'' ``Average Wind Speed,'' ``Maximum Sustained Wind Speed,'' ``Maximum Temperature,'' and ``Minimum Temperature'') and one class label (``precipitation'').
In our analysis, we predict whether rain precipitation was observed on each day.
For the prediction model, we use logistic regression implemented in Scikit-learn 0.21.3.
We randomly split the dataset into 70\% training and 30\% test data.
The training/test split is used to ensure the plausibility of the reported accuracy.

\noindent\textbf{Drift analysis}.
We conducted this case study together with E$_1$, our expert from the China Meteorological Administration. 
She works on weather prediction model development and making weather prediction decisions.
In this case, we first illustrate how DriftVis helped identify changes in the weather data distribution across seasons, and then demonstrate how it facilitated the expert in identifying abnormal weather changes long-term.
\lu{This study will show that a high drift degree could be detected when the data sees a new season, but DriftVis could also find small drifts before the drift degree reaches an alert level.}
 
\noindent\textbf{Seasonal drifts in the first year.}
The initial model, $BL_0$, was trained on data from the summer of 1949 (June 1 to August 31). 
The remainder of the data was used to simulate a streaming context, where data is reported daily. 
When E$_1$ saw a drift, she stepped in to analyze.
The drift degree first went up to 0.2 after 15 days, on Sept. 15, and the expert decided to explore the data distribution change.
In Fig.~\ref{fig:autumn_data}(a), she found that these new data were on the boundary of the existing component with relatively lower temperature, as shown in Fig.~\ref{fig:autumn_data}A and Fig.~\ref{fig:autumn_data}B.
She selected the new data and some similar data points from the historical data to build a new base learner, $BL_1$, and combined $BL_1$ with $BL_0$ to create a new ensemble prediction model.
After the adaptation, the drift went down, indicating that the adaptation was successful.\looseness=-1

\mc{Since Oct. 1, there were some data samples deviating from the blue component, and on} Oct. 23, the expert identified a new Gaussian component (yellow, as shown in Fig.~\ref{fig:autumn_data}C) in the streaming scatterplot (Fig.~\ref{fig:autumn_data}(b)). 
The expert explained, ``This is the first month of autumn, and the temperatures have begun \mc{to} decrease.
This is why a new component has emerged.''
However, she commented that the drift degree was still low, and many points in the yellow component were flipped from the blue component. 
Therefore, she decided to monitor the change closely for the following days.
As expected, the next few days' data all fell into the yellow component, and the expert noticed an increase in drift degree, which went up to 0.2 on Nov. 2.
Fig.~\ref{fig:autumn_data}(c) shows the corresponding streaming scatterplot and Fig.~\ref{fig:autumn_data}D highlights where the latest streaming data appeared. 
E$_1$ created a base learner ($BL_2$) using the yellow component data and removed $BL_1$.
She explained, ``This component does not have a full collection of autumn data yet, but in our field, we always want to adapt the model to create a better prediction as soon as we have enough data.
While we know that when more autumn data comes in, the data pattern learned using this subset may not be robust, further adaptation can always be made.''
Just as the expert explained, the drift degree stayed low for the next few days because the model learned some patterns from the yellow component, which encapsulates many properties of the incoming data.

\begin{figure}[b]
  \centering
   \vspace{-3mm}
{\includegraphics[width=\linewidth]{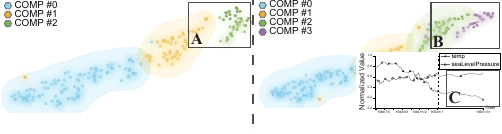}}
{\rmfamily
\put(-242,2){\textls[-5]{\spaceskip=0.16em\relax(a) A component built on Dec. 11.}}
\put(-117,2){\textls[-5]{\spaceskip=0.16em\relax(b) A component built on Jan. 31.}}
}
\vspace{-5.5mm}
  \caption{\weikairevision{The streaming scatterplot shows the distribution change in the winter of 1949:
  (a) the appearance of the green component (A);
  (b) the purple component (B) was built, which shared a similar pattern to the green component, e.g., low temperature and high sea level pressure (C).}}
  \label{fig:winter_data}
\end{figure}

On Dec. 11, a green component (Fig.~\ref{fig:winter_data}A) appeared, and on the same day, the drift degree increased rapidly to 0.15.
Similar to how she handled the occurrence of the yellow component, the expert first built a new base learner ($BL_3$) with available data in early winter and replaced it with $BL_4$ once the green component had more data and the drift degree increased.
As of Jan. 8, 1950, the ensemble model had three base learners, $BL_0$, $BL_2$, and $BL_4$.

\begin{figure}[b]
  \centering
   \vspace{-4mm}
\subfigure
{\includegraphics[width=\linewidth]{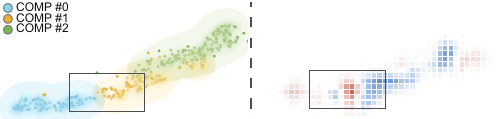}}\\
\vspace{-4mm}
  \caption{Exploring the density change of the yellow component. The left figure shows the scatterplot on May 31, 1950, and the right figure shows the density diff comparing the latest streaming data to the training data of $BL_2$ (yellow component).}
  \label{fig:first_year}
\end{figure}

In late January, the weather became very cold, and a fourth Gaussian component (purple, Fig.~\ref{fig:winter_data}B) appeared on Jan. 31, as shown in Fig.~\ref{fig:winter_data}(b).
At this time, the drift degree was not high, and 
\weikai{E$_1$ commented that these data in purple shared a similar pattern to the green component, e.g., low temperature and high sea level pressure, as illustrated in the line chart in Fig.~\ref{fig:winter_data}C}.
Based on this observation, she merged the purple component into the green component and used all of the data in the green component to create a new base learner $BL_5$ and replaced $BL_4$. 

The drift remained low for the following two months but went up again in early April, when the expert noticed a few days with relatively higher sea level pressure for the time in a year. 
The drift degree increase was a false alarm caused by a mis-clustering of the incremental GMM, and the expert corrected this by merging these points into the green component. 
By the end of the first full year, E$_1$ decided to adjust the model covering the yellow component because she saw density increased in some regions in this component.
\yafeng{Comparing the latest streaming data with the training data in $BL_2$, which was built on the yellow component earlier, Fig.~\ref{fig:first_year} shows that in the area of the yellow component, there were some red grids in the density diff mode. 
This indicated that the latest streaming data in the yellow component were likely to have a different density distribution than what had been modeled. Specifically, the area with deep red grids became denser.} 
Therefore, using a new base learner $BL_6$ trained on all data in the yellow component, E$_1$ replaced $BL_2$.
The current model ($M_1$) now includes base learner $BL_0$, $BL_6$, and $BL_5$, and the model accuracy running over the past year is 0.77.\looseness=-1

\begin{figure}[!tb]
  \centering
{\includegraphics[width=0.9\linewidth]{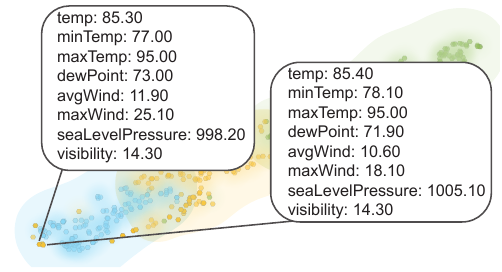}}
\vspace{-3mm}  
  \caption{The yellow points in the blue region show high max temperature and low sea level pressure and are more similar to summer data (blue component) than autumn data (yellow component). }
  \vspace{-5mm}  
  \label{fig:extreme_high_temp}
   \vspace{-1mm}
\end{figure}

\noindent\textbf{Occasional drifts after having four seasons.}
When it came to the second summer (1950), the drift degree stayed under or around 0.1 until July.
E$_1$ wanted to explore the data at the end of June because she saw some yellow points were projected in the blue density area (Fig.~\ref{fig:extreme_high_temp}).
Looking at the feature values of these yellow points, she found that these days had high max temperatures ($\sim95^{\circ}F$) and low sea level pressures ($<1006$).
She commented that these days could be extreme days in the summer when compared to the historical data.
It was more reasonable to group them into the blue component, which includes the data of last summer than associating them with the yellow component, which had typically autumn and spring data.
She merged these yellow points into the blue component and replaced $BL_0$ with a new base learner $BL_7$ using the refined blue component. 
The drift degree became lower.

After exploring the weather data for more than 14 months, the data had been relatively well covered by the model, and the drift degree stayed lower than 0.1 for the remainder of 1950 and the following year.
This model ($M_2$) comprises $BL_8$, $BL_6$, and $BL_5$. 
While more weather data streamed in, the expert conducted two more adaptations in 1952 in mid-July and mid-September, respectively, updating base learners and creating a model ($M_3$) with $BL_{9}$, $BL_{10}$, and $BL_{5}$. \looseness=-1

Comparing the precipitation prediction accuracy of the three models with their last adaptation in Jun. 1950 ($M_1$), Sept. 1950 ($M_2$), and Sept. 1952 ($M_3$), we run them on yearly test data from Jun. 1949 to May 1954. $M_1$ has its yearly accuracy being 0.77, 0.72, 0.73, 0.67, and 0.74; $M_2$ has its yearly accuracy being 0.78, 0.74, 0.73, 0.69, and 0.76; and $M_3$ has its yearly accuracy being 0.79, 0.75, 0.76, 0.73, and 0.81.
Their prediction accuracy shows that with further adaptations, $M_2$ performs better than $M_1$, and $M_3$ is even more accurate than $M_2$, especially in the fourth and fifth years. 
The improvement of the prediction accuracy also indicates that DriftVis is effective in identifying and overcoming drift.

\subsubsection{Textual Data: Academic Paper Topic Evolving}
\begin{figure*}[!tb]
  \centering
\subfigure[\yafeng{Exploring} papers from 2000 to 2003.]
{\includegraphics[width=0.44\linewidth]{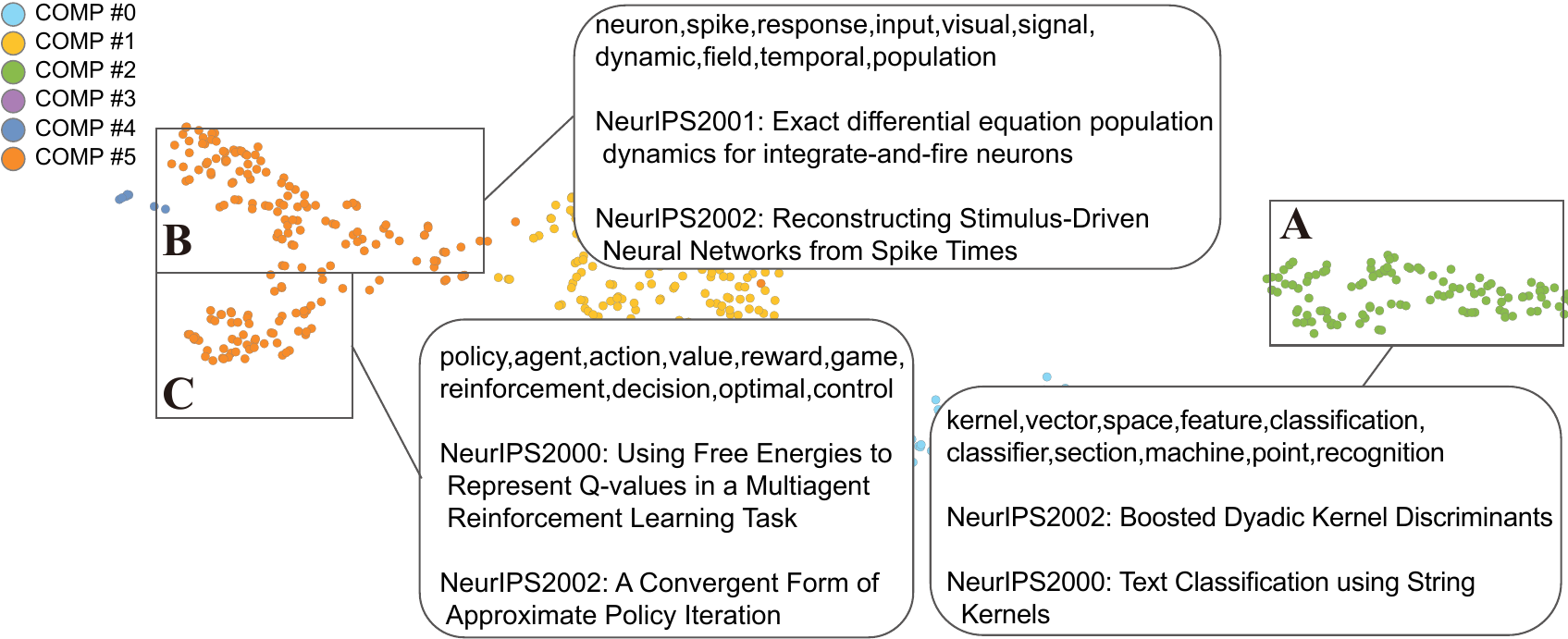}}
\hspace{0.05\linewidth}
\subfigure[\yafeng{Exploring} papers from 2004 to 2007.]
{\includegraphics[width=0.44\linewidth]{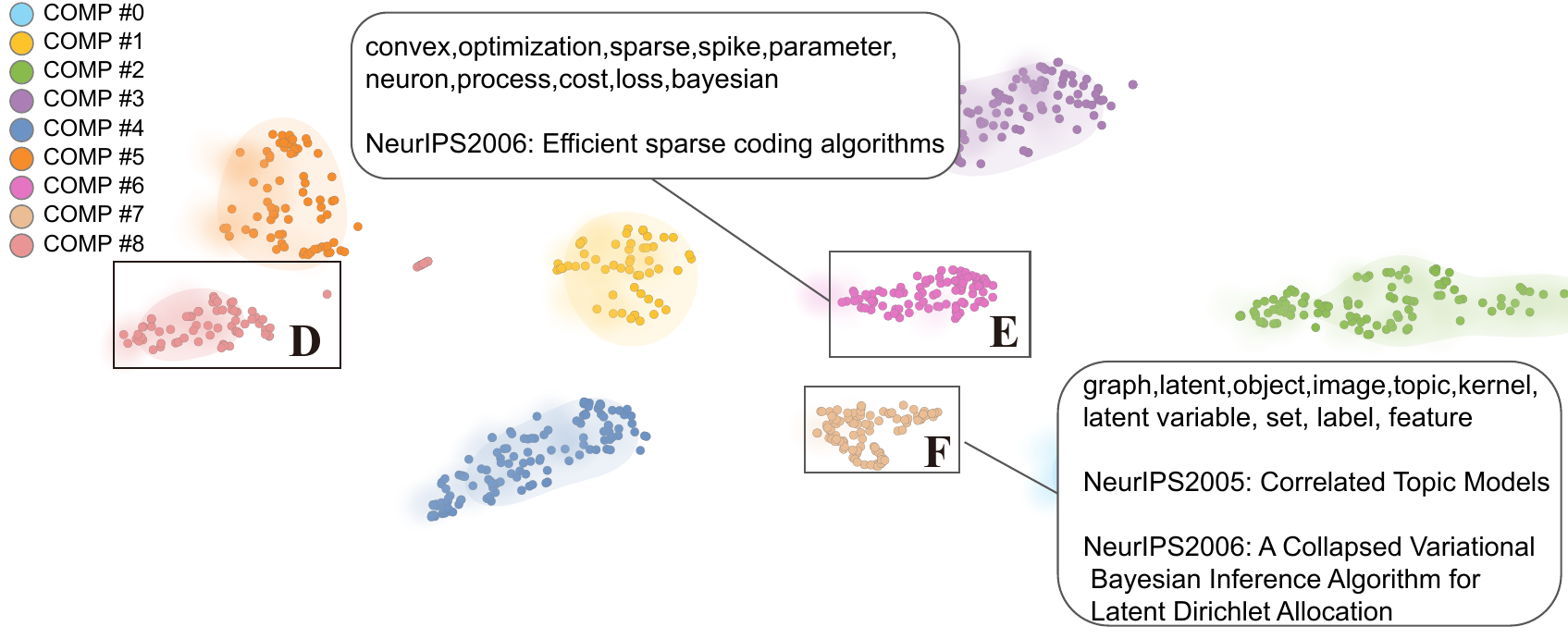}}\\
\vspace{-3mm}
\subfigure[\yafeng{Exploring} papers from 2008 to 2011.]
{\includegraphics[width=0.44\linewidth]{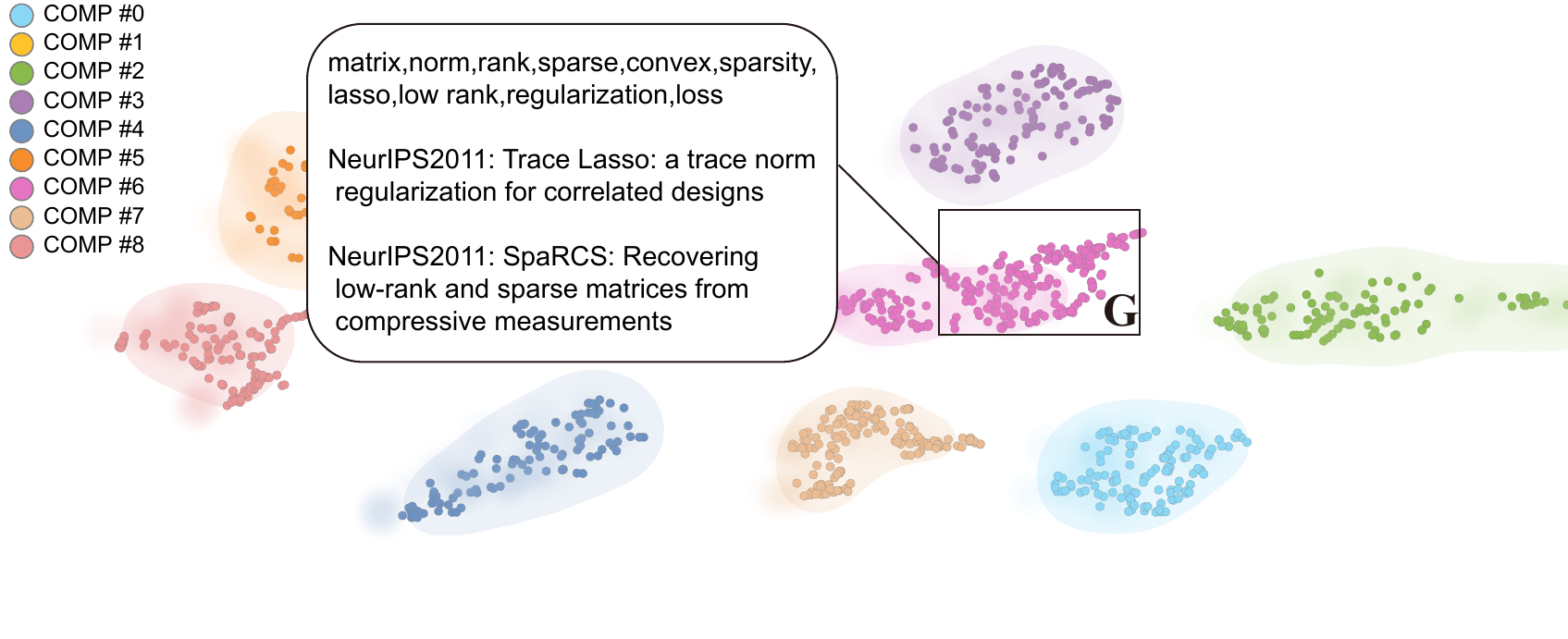}}
\hspace{0.05\linewidth}
\subfigure[\yafeng{Exploring} papers from 2016 to 2018.]
{\includegraphics[width=0.44\linewidth]{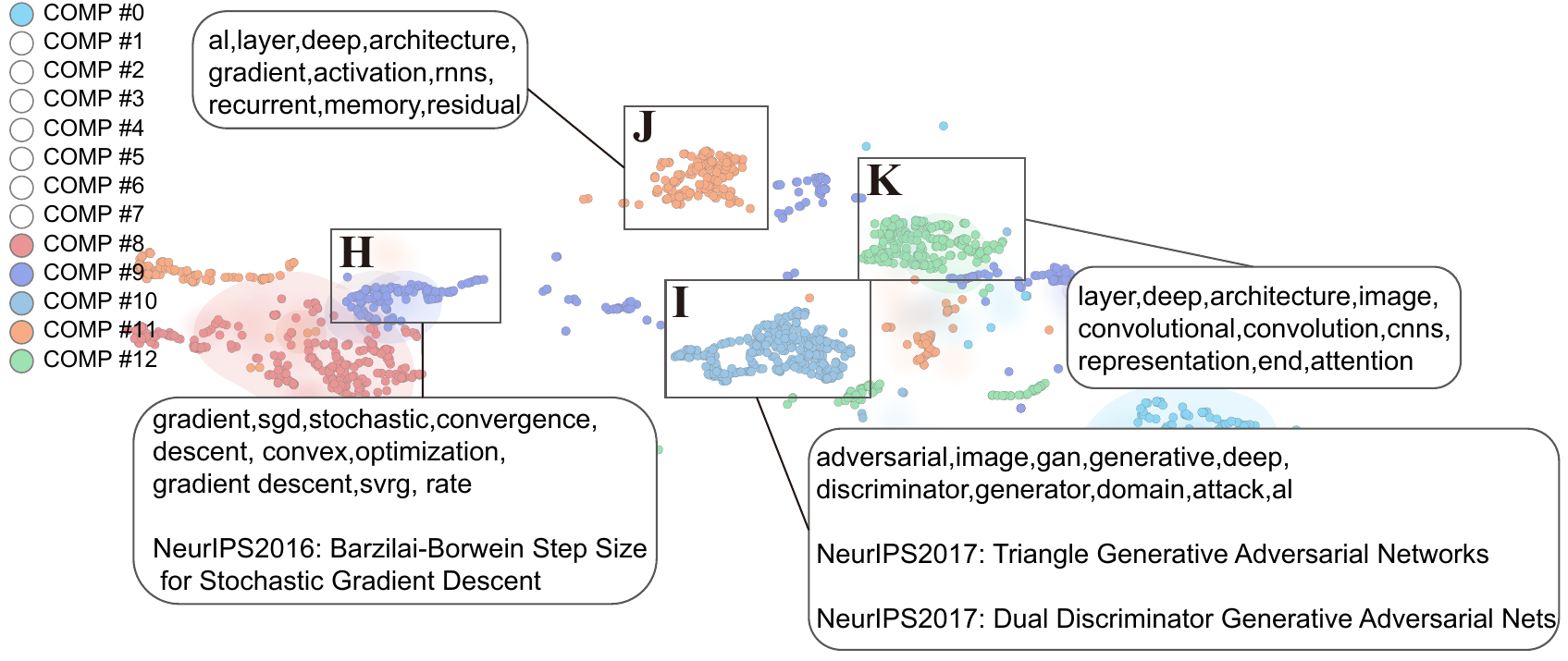}}\\
\vspace{-3mm}    
  \caption{\weikairevision{The scatterplot shows the Gaussian components in different years when analyzing NeurIPS papers. Each subgraph shows the different analytics steps our expert took. The regions (A)-(K) indicates the interesting components explored.}}
  \label{fig:text}
   \vspace{-4mm}
\end{figure*}



\noindent\textbf{Dataset.}
This dataset consists of 7,926 papers from the top-tier machine learning conference, NeurIPS (from 2000 to 2019).
For each paper, we extracted the title, abstract, and introduction and further preprocessed \yafeng{the text} by tokenization, lemmatization, keeping only the top-5000-most frequent words in the corpus. 
Each paper was then converted into a 5,000-dimension vector, and each dimension was reweighted \yafeng{using the TF-IDF scheme}~\cite{manning2008introduction}.
To accelerate the computation of the constrained t-SNE, we further reduced the dimension of each paper to $10$ by singular value decomposition (SVD).
\yafeng{The reason for using SVD is that the contribution of the words can be interpreted intuitively from each reduced dimension.}


In this case, we collaborated with E$_3$ to explore the prevalent research topics in NeurIPS.
To \mc{help E$_3$ gain insights from the past research}, we built a multi-class text classifier that predicted the topic of a paper.
To get the ground-truth label for each paper, we employed a ten-topic Latent Dirichlet Allocation~\cite{blei2003latent} to calculate the topic distribution of each paper.
Each paper's initial ground-truth label was set to the most dominant topic in the topic distribution.
We further invited E$_4$ to verify and fine-tune the ground-truth labels of ambiguous papers with mixed topics.
Given the ground-truth labels, we built an SVM-based 10-class classifier to predict the label (topic) of each paper. 
\mc{We randomly split the dataset into $60\%$ training and $40\%$ test data.
The training/test split is used to ensure the plausibility of the reported accuracy.
We used the training papers in 2000-2003 to build an initial model, whose accuracy was $0.75$.
The papers after 2003 were assumed to stream in year by year to simulate the common practice of E$_3$, where he went through the NeurIPS papers after each year's conference to discover emerging topics and trends.\looseness=-1}
 




\noindent\textbf{Analysis of papers in 2000-2003.} 
E$_3$ started the analysis with the papers in 2000-2003 (754 papers).
Fig.~\ref{fig:text}(a) shows the distribution of these papers.
Several topics were identified from this distribution by hovering over the papers in the scatterplot to examine the keywords and titles. 
For example, the keywords in the green component (Fig.\ref{fig:text}A) were ``kernel, vector, feature.''
E$_3$ commented that there was a high research interest in the kernel methods at that time due to the success of support vector machines, which was reflected in the leader-board of the MNIST dataset~\cite{MNIST_lb}.
The orange component ($\#5$) contained two inner clusters (Fig.\ref{fig:text}B and C).
The \yafeng{top} one (Fig.\ref{fig:text}B) focused on neural networks, while the bottom one (Fig.\ref{fig:text}C) focused on reinforcement learning.
E$_3$ commented that these two inner clusters might split if their respective topics become popular in the future.\looseness=-1


\noindent\textbf{Analysis of papers in 2004-2007.}
Next, the papers from 2004 to 2007 were added year by year.
In the year 2007, the drift degree increased to 0.15 and triggered E$_3$'s detailed analysis.
As shown in Fig.~\ref{fig:text}(b), he found that there were three new components compared with the training data (2000-2003).
In particular, component $\#8$ (Fig.~\ref{fig:text}D) \yafeng{separated} from the previously mixed component $\#5$ (Fig.~\ref{fig:text}B and C) because it now had a sufficient number of papers to form a new component.
The pink component (Fig.~\ref{fig:text}E) talked about convex optimization.
The keywords in the brown component (Fig.~\ref{fig:text}F) were ``latent'' and ``topic.'' 
This new component appeared due to the success of Latent Dirichlet Allocation~\cite{blei2003latent}, which was published in 2003.\looseness=-1

Since there were three new components and a relatively large drift degree of $0.15$, the expert decided to build new base learners to adapt to this drift.
Using the base learner view (Fig.~\ref{fig:teaser}D), the expert selected the papers in these three components and built three base learners with the papers in the components.
After merging these three new base learners and the old one into an ensemble model (Sec.~\ref{sec:system}),
\mc{the drift degree decreased from $0.15$ to $0.07$, and the \weikairevision{classification} accuracy improved from \textbf{0.73} to \textbf{0.75} \liu{in the papers published during 2004-2007}.\looseness=-1}

\noindent\textbf{Analysis of papers in 2012-2015.}
The drift degree increased gradually from 2012 to 2015.
In 2015, the drift degree increased to 0.14, and it prompted
E$_3$ to perform an analysis of this drift.
This drift was expected because E$_3$ knew that these years were the dawn of the deep learning era, and new machine learning approaches were developed.
He found that no new component was formed, and there were no deviating data samples.
Thus, E$_3$ switched to the density diff mode to further examine the root cause of the drift.
He found that 
the most noticeable drift happened in the pink component ($\#6$), where the density of three regions increased (Fig.~\ref{fig:teaser}G, H, and I).
The left-top part of the pink component (Fig.~\ref{fig:teaser}G) was about deep neural networks, which was a rising topic at that time.
E$_3$ commented that this new topic was triggered by AlexNet~\cite{krizhevsky2012imagenet} being the winner of the ILSVRC2012 image classification challenge.
The bottom part of the pink component (Fig.~\ref{fig:teaser}H) was about variational inference.
This topic became popular due to the introduction of an efficient deep generative model - Variational Auto-Encoder (VAE)~\cite{kingma2013auto} in 2013.
The right part of the pink component (Fig.~\ref{fig:teaser}I) was still about ``sparse,'' but ``tensor'' became a new keyword due to the increasing research interest in image data, which can be seen as a 3D tensor.
Although there were no new components, the large change of the keywords in the pink component still \yafeng{prompted} E$_3$ to adapt the model.
Since there was a noticeable drift within the pink component ($\#6$), he selected \mc{the} papers from the pink component and built a base learner with these papers.
\weikai{After the adaptation, the drift degree decreased \mc{from 0.14} to 0.08, and the performance view (Fig.~\ref{fig:teaser}F) showed that the \weikairevision{classification} accuracy increased from \textbf{0.58} to \textbf{0.74}.
Many papers about optimization (class 5) and sparse matrix (class 8) were now correctly classified.
}

\noindent\textbf{Analysis of papers in 2016-2019.}
As interest in machine learning increased, so did the volume of research output, leading to a large degree of drift in the model from the years 2016-2018, with the drift reaching 0.18.
After hiding \mc{some unchanged components}, the expert identified four new components.
All four new components were about deep-learning-related topics.
\weikai{
Component $\#9$ (Fig.~\ref{fig:text}H) was about optimization techniques for deep learning, with keywords ``gradient'' and ``sgd.''
Component $\#10$ (Fig.~\ref{fig:text}I) was about the generative adversarial network (GAN), which was first proposed in NeurIPS 2014 and quickly received much attention.
Components $\#11$ (Fig.~\ref{fig:text}J) and $\#12$ (Fig.~\ref{fig:text}K) were about recurrent neural networks (RNN) and convolutional neural networks (CNN), respectively.
These two types of deep neural networks showed great success in speech and image processing.
}
\weikai{To adapt to this large change of the data distribution, E$_3$ built a learner with the papers in the four deep-learning-related components and added it to the model.}
After the adaptation incorporating these two new learners, the \weikairevision{classification} accuracy improved from \textbf{0.59} to \textbf{0.73}, and the drift degree decreased to 0.08.

The drift degree remained nearly unchanged in 2019.
At the end of the case study, E3 commented that ``the trial of the system was really valuable to me not only because it solved the performance issue caused by concept drift, but also enabled me to take a wonderful tour of the past in machine learning.''

\section{Discussion and Future Work}
\label{sec:discussion}
Although the case studies on real-world datasets \yafeng{have demonstrated} the usefulness of DrifVis, there are several limitations, which may serve as promising avenues for future research.

\noindent \textbf{Generalization.}
In our current prototype, DriftVis has only been run on classification tasks and did not factor in known prior knowledge, if any, of the data distribution. The proposed method can be easily generalized into other tasks or to leverage known distributions, but some additional work has to be done.
To extend to other tasks, \yafeng{such as regression or clustering,}
we need to replace the performance view \yafeng{with task-specific designs, e.g., RegressionExplorer~\cite{dingen2018regressionexplorer} for regression tasks}. 
As for the data, currently, we use an incremental GMM to model the distribution of data since it is capable of approximating any continuous distributions~\cite{bishop2006pattern}.
\yafeng{Sometimes}, the analyst \yafeng{may} have \yafeng{some} prior knowledge of the data distribution, such as using a Poisson distribution \yafeng{to better} model the accumulated event number, instead of using Gaussian distribution.
To incorporate such prior knowledge, we may consider using distribution-dependent distance functions~\cite{kifer2004detecting} in drift detection.

\noindent \textbf{Drift From Other Causes.}
In DriftVis, we focus on drift that is caused by changes in the data distribution $P(X)$, where $X$ is the data.
However, there can be concept drift due to changes in label $P(Y \mid X)$ (where $Y$ are the labels) while the data distribution $P(X)$ remains unchanged.
This is called label drift.
\mcr{For example, in the image classification task, attackers can construct adversarial examples~\cite{yuan2019adversarial} to mislead the classifier.
These adversarial examples share a similar distribution with the training data, but their label is different.}
Such drift can not be detected until the ground-truth labels are obtained.
Considering the high cost of acquiring ground-truth labels, our method focuses on the data distribution change, and label drift detection is left for future work. 
To enable the support of label drift detection in DriftVis, 
two modifications are needed.
First, we can replace the distribution-based concept drift detection approach with an error-rate-based one, which relies on ground-truth labels.
Second, we can change the streaming scatterplot from visualizing $P(X)$ to showing $P(X,Y) = P(X) P(Y\mid X)$, which simultaneously covers the drift caused by $P(X)$ and  $P(Y\mid X)$.



\noindent \textbf{Scalability.}
In the weather prediction case study, we demonstrated the potential of DriftVis to handle streaming data containing thousands of time points.
\weikairevision{
When there are 1,500 historical time points, the system has a latency of about 5s to process the new time point on a PC with Intel i7-9700K CPU (3.60 GHz) and 32 GB memory.}
In practice, the \mcr{incoming} data may arrive per hour, per minute, or even per second. 
As a large amount of data samples stream in rapidly, it is challenging to update the incremental GMM and the dynamic t-SNE projection fast enough for every time point, and the potentially high number of Gaussian components can make the scatterplot hard to read.
To speed up the calculation, a distribution-based sampling approach~\cite{palmer2000density} can be used. To keep the streaming scatterplot readable,
stale or less active components can be hidden until they are activated by a certain number of new data samples.
\weikairevision{Some incremental visualization techniques can also be employed to accelerate the visualization for large-scale data~\cite{fisher2012trust}.}

\section{Conclusion}

We have developed DriftVis, a visual analytics method to support the detection, examination, and correction of concept drift in streaming data. 
As part of DriftVis, we have employed an incremental GMM in the distribution-based drift detection to efficiently remove false alarms, which can be automatically addressed by the ensemble model. 
In addition to improvements in drift detection, DriftVis also introduces the streaming scatterplot visualization that uses a GMM-based constrained t-SNE and two visual modes (scatterplot and density diff) to explain different kinds of drift in the streaming data.  
DriftVis further combines drift detection, exploration, and adaptation via coordinated interactions.
\weikairevision{A quantitative experiment and two case studies have demonstrated the usefulness of our method.}

\acknowledgments{
This research was funded by the National Key R\&D Program of
China (No.s 2018YFB1004300, 2019YFB1405703), the National Natural Science
Foundation of China (No.s 61761136020, 61672307, 61672308, 61872389), and TC190A4DA/3.
Work by Maciejewski was partially sponsored by the U.S. National Science Foundation award number 1939725.}

\small
\bibliographystyle{abbrv}
\bibliography{reference}


\end{document}